%% file: SIDA_Arxiv.tex
\documentclass[sigconf, screen, nonacm]{acmart}

\AtBeginDocument{%
  }

\setcopyright{acmlicensed}
\copyrightyear{2025}
\acmYear{2025}
\acmDOI{XXXXXXX.XXXXXXX}
\acmConference[MM'25]{the 33rd ACM International Conference on Multimedia}{Oct 27--31,
  2025}{Dublin, Ireland}

\acmISBN{978-1-4503-XXXX-X/2018/06}

\usepackage{xspace}
\usepackage{multirow}
\usepackage{amsmath} 
\usepackage{makecell}
\usepackage[linesnumbered,ruled,vlined]{algorithm2e} 
\SetKwInput{KwInput}{Input} 
\usepackage{algpseudocode}

\usepackage{adjustbox}

\newcommand{\model}{\textsl{SIDA}}

\newcommand{\Tref}[1]{Table~\ref{#1}}

\newcommand{\Fref}[1]{Fig.~\ref{#1}}

\usepackage{amsmath}
\usepackage[linesnumbered,ruled,vlined]{algorithm2e}
\usepackage{algpseudocode}
\usepackage{xcolor} 
\definecolor{mainlinecolor}{rgb}{0.0118,0.6471,0.9882}

\begin{document}

\title{SIDA: Synthetic Image Driven Zero-shot Domain Adaptation}

\author{Ye-Chan Kim}
\email{dpcksdl78@hanyang.ac.kr}
\affiliation{
  \institution{Hanyang University}
  \country{}
}

\author{SeungJu Cha}
\email{sju9020@hanyang.ac.kr}
\affiliation{
  \institution{Hanyang University}
  \country{}
}

\author{Si-Woo Kim}
\email{boreng0817@hanyang.ac.kr}
\affiliation{
  \institution{Hanyang University}
  \country{}
}

\author{Taewhan Kim}
\email{taewhan@hanyang.ac.kr}
\affiliation{
  \institution{Hanyang University}
  \country{}
}

\author{Dong-Jin Kim\textsuperscript{\dag}}
\authornote{Corresponding author.}
\email{djdkim@hanyang.ac.kr}
\affiliation{
  \institution{Hanyang University}
  \country{}
}




\input{Sec/01_abstract}

\maketitle

\input{Sec/02_intro}
\input{Sec/03_related}
\input{Sec/04_method}
\input{Sec/05_experiments}

\input{Sec/06_conclusion}
%

\begin{acks}
  This was partly supported by the Institute of Information \& Communications Technology Planning \& Evaluation (IITP) grant funded by the Korean government(MSIT) (No.RS-2020-II201373, Artificial Intelligence Graduate School Program(Hanyang University)) and the Institute of Information \&Communications Technology Planning \& Evaluation (IITP) grant funded by the Korean government(MSIT) (No.RS-2025-02219062, Self-training framework for VLM-based defect detection and explanation model in manufacturing process).
\end{acks}
\bibliographystyle{ACM-Reference-Format}
\bibliography{sample-base}

\end{document}

%% file: Sec/01_abstract.tex
\begin{abstract}
Zero-shot domain adaptation is a method for adapting a model to a target domain without utilizing target domain image data.
To enable adaptation without target images, existing studies utilize CLIP's embedding space and text description to simulate target-like style features. 
Despite the previous achievements in zero-shot domain adaptation, we observe that these text-driven methods struggle to capture complex real-world variations and significantly increase adaptation time due to their alignment process.
Instead of relying on text descriptions, we explore solutions leveraging image data, which provides diverse and more fine-grained style cues.
In this work, we propose $\model$, a novel and efficient zero-shot domain adaptation method leveraging synthetic images.
To generate synthetic images, we first create detailed, source-like images and apply image translation to reflect the style of the target domain.
We then utilize the style features of these synthetic images as a proxy for the target domain.
Based on these features, we introduce Domain Mix and Patch Style Transfer modules, which enable effective modeling of real-world variations.
In particular, Domain Mix blends multiple styles to expand the intra-domain representations, and Patch Style Transfer assigns different styles to individual patches.
We demonstrate the effectiveness of our method by showing state-of-the-art performance in diverse zero-shot adaptation scenarios, particularly in challenging domains.
Moreover, our approach achieves high efficiency by significantly reducing the overall adaptation time.
\end{abstract}

%% file: Sec/02_intro.tex
\renewcommand{\shortauthors}{Kim et al.}
\section{Introduction}
\label{sec:intro}
To tackle the inherent issue of domain shift in deep learning, various Domain Adaptation (DA) studies~\cite{ganin2016domain,shin2021labor,hoffman2018cycada} based on Unsupervised Domain Adaptation (UDA) have been proposed~\cite{hoyer2022daformer,zhang2021prototypical,zou2018unsupervised,zhu2023patch}.
However, these approaches often struggle to adapt in dangerous or infrequent scenarios (e.g., ``fire'' and ``sandstorms'') due to the limited number of target domain image data~\cite{fahes2023poda,yang2024unified}.
To address this limitation, 
Zero-Shot Domain Adaptation (ZSDA), which does not rely on any target domain image data, has emerged~\cite{fahes2023poda,yang2024unified}.
However, the absence of target images introduces considerable challenges in effective domain adaptation.
As a result, most existing approaches rely on textual descriptions to approximate target domain characteristics.
For example, P{\O}DA\xspace~\cite{fahes2023poda} and ULDA~\cite{yang2024unified} utilize CLIP's~\cite{radford2021learning} language-embedding capabilities to optimize learnable vectors in the style of the target domain.
They then apply the style transfer algorithm~\cite{huang2017arbitrary} with optimized vectors in feature space to obtain target domain-like features without target domain image data.
\begin{figure}[t]
    \centering
    \includegraphics[width=\linewidth]{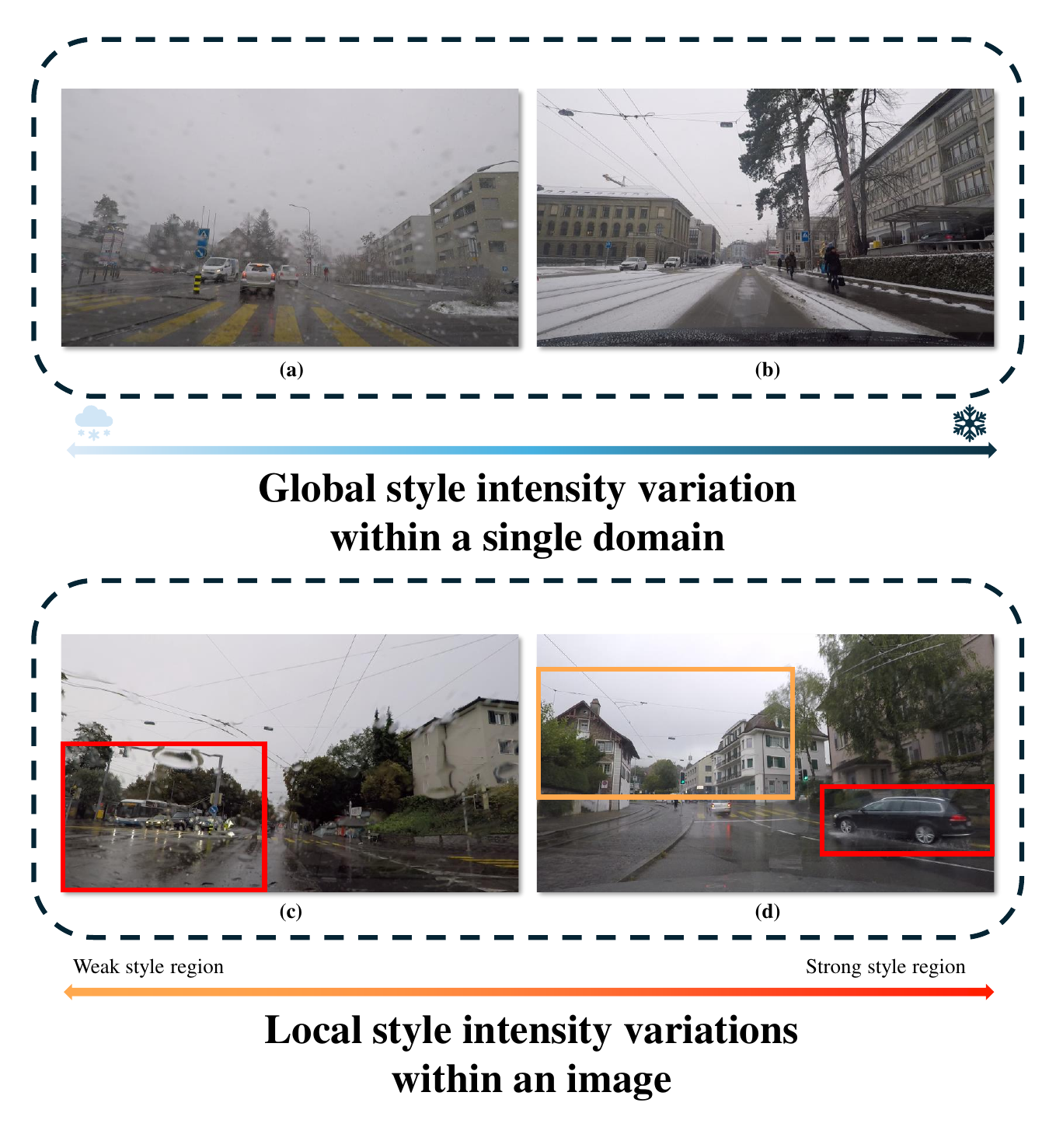}
    \caption{
    \textbf{Global and local style intensity variations in real-world.}
   Images (a) and (b) are both from the snow domain.
   These images illustrate variations in global style intensity.
   While (b) clearly depicts a snowy scene, (a) appears to resemble a rainy one. Images (c) and (d), from the rain domain, demonstrate local style intensity variations. Red regions highlight strong rain characteristics, whereas orange regions indicate weaker rain-related styles.
}
    \label{fig:example_tmp}
\end{figure}

However, despite the improvements of P{\O}DA\xspace~\cite{fahes2023poda} and ULDA~\cite{yang2024unified}, several key limitations remain in leveraging text descriptions to make target-like style features.
First, the simple fixed text descriptions (e.g., {``Driving at \{domain\}.''}) to simulate target style are not diverse enough to reflect real-world environments.
In Fig.~\ref{fig:example_tmp}, we show variations in global style intensity within the same domain ({(a) and (b)}) and local style variation in a single image ({(c) and (d)}).
Despite these variations, existing methods overlook the differences in style intensity that are often observed in real-world scenarios.
They typically rely on fixed text descriptions and a single simulated target-like style feature.
Second, text-driven methods require an additional alignment process for each source image, which aims to optimize learnable vectors applied to image features.
This alignment process becomes inefficient as the source data size grows.

To overcome these limitations, we aim to derive target-like style features in the \emph{image domain}.
We note the ability of recent generative models~\cite{rombach2022high,ramesh2021zero,cha2025verbdiff,Oh2025catch} to effectively generate diverse and real-world-like images.
Building on this, we propose \textbf{$\model$} (\textbf{S}ynthetic \textbf{I}mage Driven Zero-shot \textbf{D}omain \textbf{A}daptation), a novel and efficient zero-shot domain adaptation method that effectively leverages the target-like style features from \emph{synthetic images}.
The architecture of $\model$ comprises three key stages: Image generation process, Domain Mix \& Patch Style Transfer module, and fine-tuning stage.
First, we extract scene descriptions from source images via a VLM-based caption generator~\cite{achiam2023gpt}. Given the extracted descriptions, we generate various source-like synthetic images mirroring the contents of the real source image via an image generator~\cite{rombach2022high}. 
After that, we translate the source-like synthetic images into various target domains using an image translation model~\cite{rombach2022high}.

Then, we introduce (1) Domain Mix and (2) Patch Style Transfer to enrich global and local style variations.
Domain Mix is designed to obtain diverse target-like styles by simulating variations in global style intensity.
Specifically, we blend the style features of synthesized images from the main target domain and an auxiliary domain.
Additionally, to help the model capture the variability in local style intensity, we introduce Patch Style Transfer, which applies distinct target-like style features to each patch.
We utilize the diverse target-like styles generated by Domain Mix with varying intensities on a per-patch basis.
Lastly, we fine-tune the model by introducing a weighted cross-entropy loss function, based on entropy information, to effectively learn appropriate weights for the novel style.

We validate $\model$ across various existing ZSDA experimental settings.
Our method consistently shows superior performance and efficiency in various scenarios.
In particular, $\model$ shows improved performance in challenging environments, such as fire and sandstorms, where ZSDA is most crucial due to the scarcity of data.

In summary, our contributions are as follows:
\begin{itemize}
\item We propose $\model$, an efficient and effective ZSDA method that leverages synthetic images from the image domain instead of relying on text descriptions.
\item We propose Domain Mix and Patch Style Transfer, which simulate a wide range of diverse global style intensities and local style variations like real-world scenarios.
\item Extensive experiments show that $\model$ with additional entropy information achieves meaningful performance improvements in various ZSDA settings.
\end{itemize}

%% file: Sec/03_related.tex
\renewcommand{\shortauthors}{Kim et al.}
\section{Related Work}
\label{sec:related}

\noindent\textbf{Zero-shot domain adaptation.}
Unlike UDA and Domain Generalization (DG)~\cite{qiao2020learning,zhou2021domain} that rely on target domain image data, 
recent ZSDA approaches~\cite{fahes2023poda,yang2024unified} have utilized CLIP embedding~\cite{radford2021learning} to generate target-adaptive features through text-driven style transformations.
P{\O}DA\xspace~\cite{fahes2023poda} introduced a PIN module to transform source domain features into target-like representations.
This module employs a learnable vector as a proxy for the target style and optimizes these vectors to align with the target text description feature for each individual source image.
Once the optimization completes, P{\O}DA\xspace~\cite{fahes2023poda} applies style transfer~\cite{huang2017arbitrary} to the source image features using the optimized vector to generate target-like stylized features. 
They then fine-tune the model on these stylized features, enabling it to adapt to the target domain.
ULDA~\cite{yang2024unified} extends P{\O}DA\xspace to enable multi-target domain adaptation.
They optimize learnable vectors hierarchically and allow a single model to adapt across multiple target domains without requiring separate domain-specific models. 
While ULDA improves zero-shot adaptability, both P{\O}DA\xspace and ULDA require individual optimization processes for each source image, making them time-consuming and heavily dependent on the size of the source dataset. 
This reliance on per-image optimization introduces scalability challenges, particularly when applied to large-scale source datasets.
Moreover, these methods rely solely on fixed text descriptions for learnable vector optimization, such as ``Driving at \{domain\},'' which limits their ability to capture the diverse characteristics of the real world.
In contrast, we propose using synthetic images instead of simple text descriptions to capture the diverse visual information of the target domain.

\noindent\textbf{Synthetic image adaptation.}
As the quality of synthetic image generation reaches photo-realistic, various studies have been conducted about utilizing synthetic images~\cite{Kim2025sync}.
In domain adaptation,~\cite{benigmim2023one} fine-tunes a DreamBooth~\cite{ruiz2023dreambooth} to generate a large volume of pseudo-target domain images.
After the data generation stage, they adapt the model using the UDA method~\cite{hoyer2022daformer}.
\cite{shen2024controluda} proposed enhancing UDA performance under adverse weather conditions by using synthetic images.
They generate synthetic images by fine-tuning a diffusion model utilizing pseudo-labels derived from a pre-trained UDA model and structural information obtained through an edge detector.
In domain generalization,~\cite{niemeijer2024generalization} leverages the diffusion model~\cite{rombach2022high} with text prompt tuning to make pseudo-target domain datasets.
\cite{qian2024weatherdg} utilizes a Large Language Model (LLM) to generate prompts for fine-tuning image generators, subsequently constructing synthetic domain datasets.
However, such existing methods require an additional process of fine-tuning the image generator and constructing a large-scale external synthetic dataset, which not only increases computational cost but also degrades overall training efficiency.
In contrast, our approach does not require any fine-tuning image generator and works effectively even with only a small number of synthetic images.
\begin{figure*}[!tb]
    \centering
    \includegraphics[width=\textwidth,height=8cm]{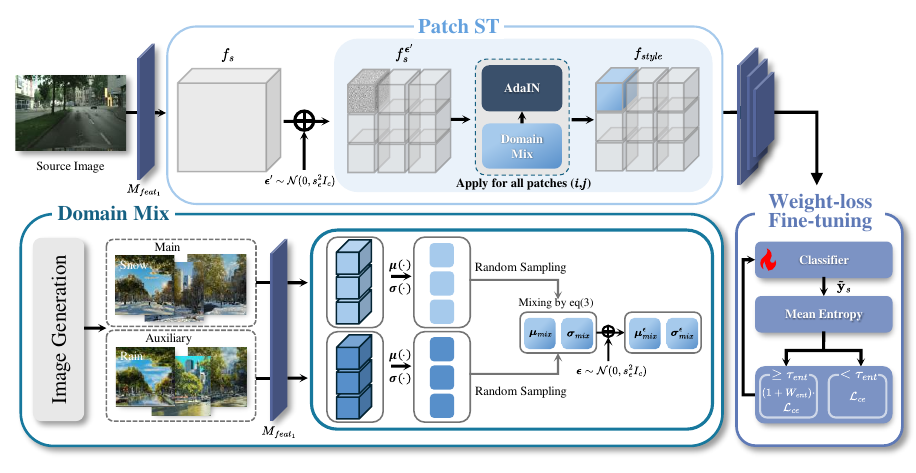}
    \caption{
    \textbf{Pipeline of $\model$.}
    Our method applies Domain Mix to the synthetic image features generated for each domain, producing diverse target-like styles. Patch Style Transfer then leverages these features to apply locally varying styles to each patch, enabling fine-grained style diversity.
    }
    \label{fig:method}
\end{figure*}

%% file: Sec/04_method.tex
\renewcommand{\shortauthors}{Kim et al.}
\section{Method}
\label{sec:method}
Our objective is to improve the performance of the pre-trained segmentation model {$M(\cdot)$} on a source domain dataset 
$\mathcal{D}_s=\{(\mathbf{x}_s,\mathbf{y}_s)\}$, when applied to an unseen target {test} dataset $\mathcal{D}_t=\{\mathbf{x}_t\}$ of the domain $t\in \mathcal{T}$, where $\mathcal{T}$ is the set of target domains
(e.g., $\mathcal{T}$=\{``night,'' ``snow,''\ldots\}).  
The model consists of two components: the feature extractor $M_{feat}(\cdot)$ and the classifier $M_{cls}(\cdot)$.
The low-level feature  $f$, used for feature transfer, is obtained by $f = M_{{feat_1}}(\mathbf{x})\in\mathbb{R}^{h \times w \times c}$.

In this section, we describe how $\model$ generates and utilizes synthetic images in ZSDA. 
Rather than relying on fixed text descriptions, our approach aims to better capture the diverse global and local style intensities observed in the real world by utilizing synthetic images.
First, we propose an image generation framework that produces target domain-like images with detailed scene components.
After generating the synthetic images, we employ Domain Mix and Patch Style Transfer to the features of these images.
Domain Mix and Patch Style Transfer simulate a wide range of target domain style features from a few synthetic images.
In the fine-tuning stage, we utilize weight loss leveraging entropy information.
The overall process of the proposed method is illustrated in~\Fref {fig:method}.

\subsection{Image Generation Process}\label{method3.1}
To obtain target domain-like images, we first generate a small set of images that closely resemble the scene composition of the source images using Stable Diffusion (SD)~\cite{rombach2022high}.
However, a simple prompt like ``Driving at \{domain\}.'' for SD is insufficient, as it fails to generate the diverse set of objects present in the source domain image as shown in~\Fref{fig:genmethod_qual}.
\begin{figure}
    \centering
    \includegraphics[width=\linewidth]{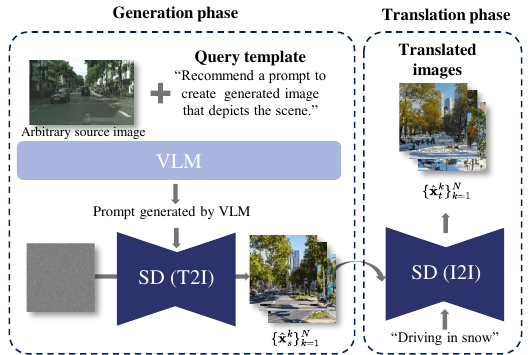}
    \caption{
    \textbf{Illustration of image generation process.}
    In the image generation process, we generate source-like synthetic images and apply a translation process to incorporate the characteristics of the target domain style.}  
    \label{fig:imagegen}
\end{figure}
Furthermore, manually crafting prompts to capture the intricate details of cityscape scenes in source images is labor intensive.
Given that refining prompts through LLM improves the quality of generated images~\cite{hao2024optimizing}, we extract detailed scene descriptions from the source image using a Vision Language Model (VLM)~\cite{achiam2023gpt}.
Specifically, we provide VLM with a structured query template (e.g., ``Recommend a prompt to create a generated image that depicts the scene.'') alongside an arbitrary source image.
With these scene descriptions, we generate a total of $N$ source-like images $\{\hat{\mathbf{x}}_s^k\}_{k=1}^N$  that effectively incorporate complex and diverse objects within the scene through SD.

Then, to generate target-like images with the characteristics of the target domain, we employ an image translation process with SD.
To synthesize images that reflect the target domain style, we feed source-like synthetic images into SD along with the prompt ``Driving at \{domain\}.''
By translating the source-like synthetic images into target domain-like style, we obtain images that preserve the semantics of the source while incorporating the domain-specific features.
We generate $N$ synthetic translated images, $\{\hat{\mathbf{x}}_t^k\}_{k=1}^N$, for each target domain $t$.
Following the image generation process, we apply two modules—Domain Mix and Patch Style Transfer—to effectively leverage the target domain style features based on the $N$ synthetic translated images for each domain $t$.

\subsection{Domain Mix \& Patch Style Transfer}\label{method3.2}
Although the translated images accurately capture the target domain style, they exhibit subtle differences compared to real images.
Unlike the diverse style intensity variations found in real-world images, the translated images display a consistent style intensity across samples and within the local regions of each image.
To reflect real-world style variations, we propose Domain Mix for capturing diverse global style intensities and Patch Style Transfer for representing local variations across image regions.

\noindent\textbf{Domain Mix.}
To capture global style variations observed in real-world settings, as illustrated in \Fref{fig:example_tmp}(a) and (b), we focus on representing diverse target style intensities within a single domain.
Specifically, we simulate diverse style features within the target domain by blending the style features of the main domain with those from an auxiliary domain.
For instance, in adaptation to the “snow” domain, we designate ``snow'' as the main domain and utilize one of the ``non-snow'' domains (e.g., ``night,'' ``rain,''\ldots) as an auxiliary domain.
First, a main target domain feature, denoted as $f_{\mathcal{M}}$, is obtained by passing a main domain sample through the first layer of the feature extractor, i.e., $f{_\mathcal{M}} = M_{{feat}_1}(\hat{\mathbf{x}}{_{snow}})$.
To select the corresponding auxiliary domain for $f_{\mathcal{M}}$, we compute the cosine similarity between $f_{\mathcal{M}}$ and each extracted feature from the other domains.
Then, we select the domain exhibiting the highest cosine similarity with $f_{\mathcal{M}}$ as the auxiliary domain $\mathcal{A}$:
\begin{equation}
\mathcal{A} = \underset{t \in \mathcal{T} \setminus \mathcal{M}}{\operatorname{argmax}}\left(\underset{k \in \{1,N\}}{\operatorname{\max}}\left( \cos\left(f_{\mathcal{M}}, f_{t}^k\right) \right)\right),
\end{equation}
where $f_t^k = M_{feat_1}(\hat{\mathbf{x}}_t^k)$.
Once the main and auxiliary domains are determined, we calculate channel-wise mean and standard deviation from each feature, $f_{\mathcal{M}}$ and $f_{\mathcal{A}}$.
$\mu(\cdot)$ and $\sigma(\cdot)$ 
denote the functions returning the channel-wise statistics of a feature:
\begin{equation}
\begin{aligned}
\boldsymbol{\mu}_\mathcal{M} = \mu(f_{\mathcal{M}}), \quad \boldsymbol{\mu}_\mathcal{A} = \mu(f_{\mathcal{A}}),
\\
\boldsymbol{\sigma}_\mathcal{M} = \sigma(f_\mathcal{M}), \quad \boldsymbol{\sigma}_\mathcal{A} = \sigma(f_\mathcal{A}).
\end{aligned}
\label{channelwise}
\end{equation} 
Subsequently, we blend them using element-wise multiplication to obtain the domain mix style feature vector $(\boldsymbol{\mu}_{mix},\boldsymbol{\sigma}_{mix})$: 
\begin{equation}
\begin{aligned}
\boldsymbol{\mu}_{mix}(\boldsymbol\lambda) = \boldsymbol\lambda\circ\boldsymbol{\mu}_{\mathcal{M}} +( \boldsymbol{1}_c-\boldsymbol\lambda)\circ\boldsymbol{\mu}_{{\mathcal{A}}},
\\
\boldsymbol{\sigma}_{mix}(\boldsymbol\lambda) = \boldsymbol\lambda\circ\boldsymbol{\sigma}_{\mathcal{M}} +( \boldsymbol{1}_c-\boldsymbol\lambda)\circ\boldsymbol{\sigma}_{{\mathcal{A}}},
\end{aligned}
\label{doaminmix}
\end{equation} 
where $\boldsymbol{1}_c \in \mathbb{R}^c$ is a vector where all elements equal to 1.
$\boldsymbol\lambda\in\mathbb{R}^c$ is a vector of per-channel mixing weights uniformly sampled from $[0, 1]^c$.
Moreover, to better reflect the diverse and complex variations even within similar styles, we add Gaussian noise $\boldsymbol\epsilon \sim \mathcal{N}(0,s_e^2I_c)$ into the $(\boldsymbol{\mu}_{mix},\boldsymbol{\sigma}_{mix})$ similar to~\cite{lee2024ifcap,fan2023towards,kim2025vipcap,su2024domain}.
Finally, we obtain augmented  $\boldsymbol{\mu}_{mix}^{\boldsymbol{\epsilon}}, \boldsymbol{\sigma}_{mix}^{\boldsymbol{\epsilon}}$ as follows:
\begin{equation}
\begin{aligned}
    \boldsymbol{\mu}_{mix}^{\boldsymbol\epsilon}( \boldsymbol\lambda,\boldsymbol\epsilon)=\boldsymbol{\mu}_{mix}(\boldsymbol\lambda)+\boldsymbol\epsilon,
    \\
    \boldsymbol{\sigma}_{mix}^{\boldsymbol\epsilon}(\boldsymbol\lambda, \boldsymbol\epsilon)=
    \boldsymbol{\sigma}_{mix}(\boldsymbol\lambda)+\boldsymbol\epsilon.
\end{aligned}
\label{equation:mix_noise}
\end{equation}
By utilizing Domain Mix, we are able to generate diverse target-like style features from only a few synthetic images and effectively reflect the diversity of global style in the real world. 
The visualization result \Fref{fig:tSNE} demonstrates the effectiveness of Domain Mix.

\noindent\textbf{Patch Style Transfer.}
To obtain target-like stylized features, previous methods ~\cite{fahes2023poda,yang2024unified}  utilize the AdaIN algorithm~\cite{huang2017arbitrary} at the feature level.
In this process, they use a single optimized target-like style feature $(\boldsymbol{\mu}, \boldsymbol{\sigma})$ and apply it uniformly to the entire source feature:
\begin{equation}
\begin{aligned}
\text{AdaIN}\bigl(f_s,\boldsymbol{\mu},\boldsymbol{\sigma}\bigr)=\boldsymbol{\sigma} \left( \frac{f_s - \mu(f_s)}{\sigma(f_s)} \right) + \boldsymbol{\mu}.
\end{aligned}
\end{equation}
However, instead of leveraging a single target-like style feature across the entire feature map, we aim to exhibit local style difference across image regions, as illustrated in~\Fref{fig:example_tmp} (c) and (d).
In Patch Style Transfer, we patchify the feature from real source images and transfer them using multiple target-like style features, enabling a wider range of style variations at the local level.

First, we extract low-level feature $f_s$ from source domain images through the first layer of the feature extractor $f_s=M_{feat_1}(\mathbf{x}_s)$.
Then, we perturb the feature $f_s$ to enhance feature diversity in the source feature space.
We multiply the channel-wise mean $\boldsymbol\mu_s = \mu(f_s)$ and standard deviation $\boldsymbol\sigma_s = \sigma(f_s)$ of the source feature by an additional noise term $\boldsymbol\epsilon^\prime$, sampled from the same distribution as $\boldsymbol\epsilon$.
We then apply the AdaIN algorithm as in prior work~\cite{fan2023towards,su2024domain}:
\begin{equation}
\begin{aligned}
f_s^{\boldsymbol\epsilon^\prime}=\text{AdaIN}\bigl(f_s,(\boldsymbol{1}_c+\boldsymbol\epsilon^\prime)\boldsymbol\mu_s,(\boldsymbol{1}_c    +\boldsymbol\epsilon^\prime)\boldsymbol\sigma_s\bigr).
\end{aligned}
\label{equation:source_noise}
\end{equation}
Next, we divide the perturbed low-level feature $ f_s^{\boldsymbol\epsilon^\prime}$ into $m\times m$ non-overlapping patches $f_s^{{\boldsymbol\epsilon^\prime}}(i,j) \in \mathbb{R}^{\frac{h}{{m}} \times \frac{w}{{m}} \times c}$, 
where $i, j \in [1, m]$ denotes the patch index.
For each patch $f_s^{{\boldsymbol\epsilon^\prime}}(i,j)$, we determine its target-like style through the Domain Mix by applying distinct ratios and noise $(\boldsymbol\lambda^{(i,j)},\boldsymbol\epsilon^{(i,j)})$, yielding diverse styles as follows:
\begin{equation}
\begin{aligned}
\boldsymbol{\mu}^{(i,j)}=\boldsymbol{\mu}_{mix}^{\boldsymbol\epsilon}( \boldsymbol\lambda^{(i,j)},\boldsymbol\epsilon^{(i,j)}),
\\
\boldsymbol{\sigma}^{(i,j)}=\boldsymbol{\sigma}_{mix}^{\boldsymbol\epsilon}( \boldsymbol\lambda^{(i,j)},\boldsymbol\epsilon^{(i,j)}).
\end{aligned}
\label{equation:PST_target}
\end{equation}
This approach ensures that each patch exhibits subtly different target-like style features.
After determining the target-like style features $(\boldsymbol{\mu}^{(i,j)}, \boldsymbol{\sigma}^{(i,j)})$ for each patch, we transfer these features to their corresponding patch regions.
Finally, we spatially combine the $m\times m$ style-transferred patches $f_s^{\boldsymbol\epsilon’}(i,j)$ to reconstruct the full feature map $f_{{style}} \in \mathbb{R}^{h \times w \times c}$, so that it has the same resolution as the source feature $f_s$.
\begin{equation}
\begin{aligned} 
f_{style}=\{\text{AdaIN}(f_s^{\boldsymbol\epsilon^\prime}(i,j),\boldsymbol{\mu}^{(i,j)},\boldsymbol{\sigma}^{(i,j)})| i, j \in [1, m]\}.
\\
\end{aligned}
\label{equation:PST}
\end{equation}
Through this process, our target-like stylized feature $f_{style}$ effectively represents varying target-like style intensities across different regions within a single image.
\subsection{Fine-tuning Stage}\label{method3.3}
In the fine-tuning stage, we fine-tune only the classifier $M_{cls}$ with the target-like stylized feature $f_{style}$.
Previous studies fine-tune models using standard Cross-Entropy (CE) loss~\cite{chen2018encoder}, whereas our approach using weighted CE loss focuses more on learning from target-like stylized feature samples.
We hypothesize that the features $f_{style}$ after the Domain Mix and Patch Style Transfer have a different distribution from that of the source domain $f_s$.
This leads the target-style-transferred feature samples to act as uncertain samples, producing high entropy in the source pre-trained classifier~\cite{wang2023informative}.
Accordingly, we propose emphasizing uncertain samples by assigning higher training weights to them, allowing the model to better focus on learning new target styles.
First, we feed the target-like stylized feature $f_{style}$ into a pre-trained classifier $M_{cls}$ to obtain 
output probability $\tilde{\mathbf{y}}_s$.
Then, we compute the entropy information using the output probability:
\begin{equation}
\begin{aligned}
W_{ent} = \mathcal{E}(\tilde{\mathbf{y}}_s),\quad 
\tilde{\mathbf{y}}_s = M_{cls}(f_{style}), 
\end{aligned}
\label{equation:AdaIN}
\end{equation}
where $\mathcal{E}(\cdot)$ is the spatial mean of the entropy of the input probability.
We apply a weighted CE loss by adding the entropy value to the original loss when $W_{ent}$ exceeds a threshold $\tau_{ent}$, and we use the standard CE loss otherwise.
In summary, the loss weight $W$ is defined as $W=(1+W_{ent} \cdot \mathbb{I} (W_{ent} \geq \tau_{ent}))$.
We train our model with the weighted loss $\mathcal{L}_{cls}$ as follows:
\begin{equation}
\begin{aligned}
\mathcal{L}_{cls} = W \cdot\mathcal{L}_{ce}.
\end{aligned}
\end{equation}
This entropy-based weighted loss ensures that the model prioritizes learning from target-like stylized samples with high uncertainty. 

%% file: Sec/05_experiments.tex
\renewcommand{\shortauthors}{Kim et al.}
\section{Experiments}
\subsection{Experimental Setup}
\noindent\textbf{Implementation details.}
We use the same pre-trained backbone model~\cite{chen2018encoder} in previous works~\cite{fahes2023poda,yang2024unified}.
Backbone model's $M_{feat}$ initialized with pre-trained CLIP-ResNet-50 weights, we fine-tune only the $M_{cls}$ weights.
The hyperparameters are set as follows: Batch size is 8,
we use Stochastic Gradient Descent (SGD)
with a momentum of 0.9 and weight decay of $10^{-3}$ is used as an optimizer.
The total number of iterations for fine-tuning was set to 2000 with a Polynomial scheduler.
In the image generation process, we use GPT-4o~\cite{achiam2023gpt} as the VLM and SD v1.5 for image generation and translation tasks. 
The number of synthetic images per domain $N$=3.
In Domain Mix and Patch Style Transfer, the number of patches 
$m$ is 3, and both noise terms 
$\boldsymbol{\epsilon}$ and $\boldsymbol{\epsilon}^{\prime}$ 
are sampled from Gaussian distributions with the same variance $s_e^2$ of $0.075^2$.
In the fine-tuning stage, the threshold of mean entropy $\tau_{ent}$ is 1.0.

\noindent\textbf{Evaluation metric.}
Following the previous work~\cite{fahes2023poda}, we evaluate our models on the validation sets of multiple unseen target domains to assess adaptation performance.
We employ Mean Intersection Over Union (mIoU\%) as the evaluation metric.
Additionally, we measured the total training time required for adaptation to evaluate efficiency.

\noindent\textbf{Datasets.}
For source images, we use the Cityscapes(CS) dataset~\cite{cordts2016cityscapes}, which consists of 2,975 training images and 500 validation images and contains a total of 19 semantic classes.
For target domain data, we use the ACDC dataset~\cite{sakaridis2021acdc}, which includes images representing four different weather domains (night, snow, rain, and fog).
Additionally, we use the GTA5 dataset~\cite{richter2016playing} for adaptation scenarios of real-to-synthetic and synthetic-to-real.
\subsection{Main Experiments}
\input{tables/main_experiment}

Following previous studies~\cite{yang2024unified}, we compare $\model$ performance across various adaptation scenarios: day$\rightarrow$night, clear$\rightarrow$snow, clear $\rightarrow$ rain, clear$\rightarrow$fog, 
 real$\rightarrow$synthetic and synthetic$\rightarrow$real.  
We compare $\model$ with previous ZSDA methods, CLIPstyler~\cite{kwon2022clipstyler}, P{\O}DA\xspace~\cite{fahes2023poda} and ULDA~\cite{yang2024unified}.
All the experiments are conducted with five different seeds, and we report the mean and standard deviation across five results.
\Tref{tab:main} shows that $\model$ consistently outperforms text-driven zero-shot domain adaptation methods like P{\O}DA\xspace and ULDA across all adaptation scenarios.
Across all scenarios, $\model$ consistently outperforms previous methods, demonstrating its robustness and effectiveness. 
We attribute this improvement to Domain Mix and Patch Style Transfer, which effectively simulate global and local style intensity variations in real-world scenarios.
These components collectively provide a more comprehensive representation, leading to enhanced adaptation performance.
\begin{figure*}[t]
  \centering
  \includegraphics[width=\textwidth]{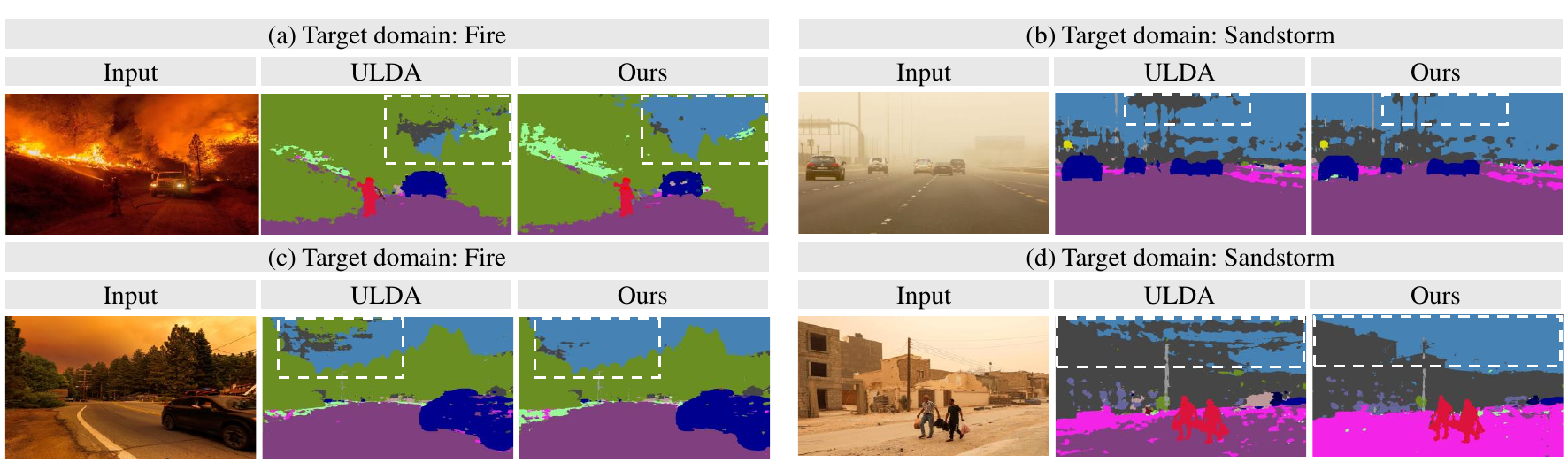}
  \caption{
     \textbf{Qualitative results on Fire and Sandstorm domains.}
    (a) and (b) show qualitative results obtained on the Sand-Fire dataset released by ULDA~\cite{yang2024unified}, while (c) and (d) illustrate the outcomes when applied to in-the-wild data.
  }
  \label{Fig:3}
\end{figure*}

\noindent\textbf{Performance in challenging domains.} We conduct experiments on challenging adaptation scenarios using the fire and sandstorm datasets constructed in ULDA~\cite{yang2024unified}. 
Following the previous experiment setting, we choose Cityscapes as the source domain and fire and sandstorm as the target domain.
The implementation details of all other models remain unchanged.
\Tref{tab:fire&sand} demonstrates that our proposed method consistently outperforms competing approaches.
Specifically, compared to ULDA, our model achieves notable improvements of 2.62\% and 2.30\% in the adaptation scenarios of CS$\rightarrow$Fire and CS$\rightarrow$Sandstorm, respectively.
This demonstrates the effectiveness of our image-based, diverse target-like features in challenging target domains, which are difficult to represent using simple textual descriptions.
We further show qualitative segmentation results in Fig.~\ref{Fig:3}.
Compared to the previous model that struggles to segment areas like the sky and buildings, $\model$ successfully captures these regions.
These results highlight the effectiveness of our method in challenging domains where ZSDA is most critical.
\input{tables/sand_fire}
\input{tables/component_ablation}
\begin{figure}
    \centering    
    \includegraphics[width=\linewidth]{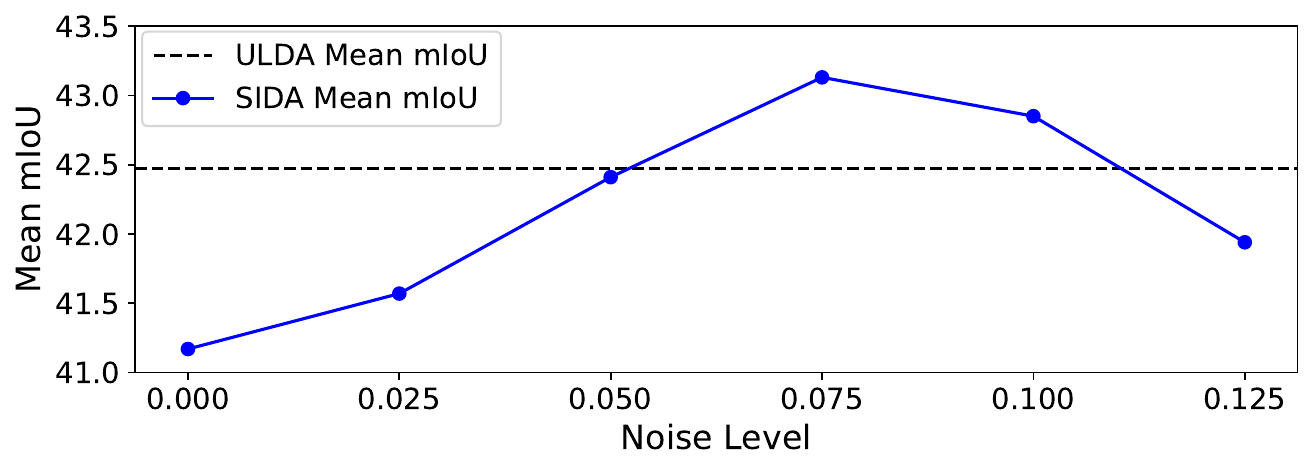}
    \caption{
    {Ablation study on noise hyperparameters of $\model$. Our method with the noise level between $s_e=0.075$ and  $s_e=0.1$ outperforms the state-of-the-art method.}
    }
    \label{fig:noise_ablation}
\end{figure}
\subsection{Ablation Study}
\begin{figure*}[t]
  \centering
  \includegraphics[width=\textwidth]{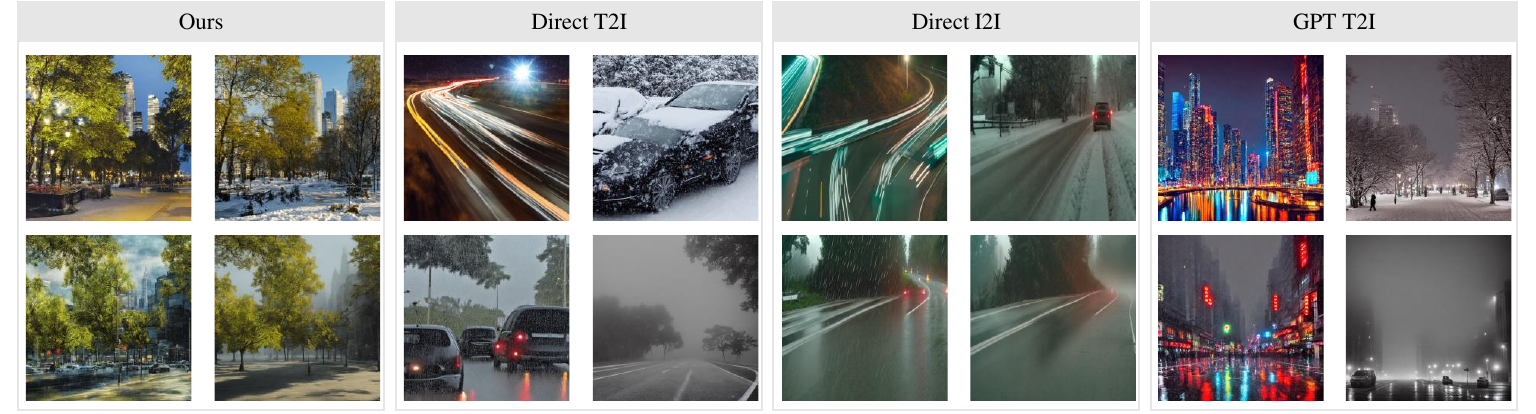}
  \caption{
     \textbf{
    Qualitative results for each generation method across four target domains—Night, Snow, Rain, and Fog (from top to bottom).}
   Our image generation method effectively captures a diverse range of target domain styles with similar object composition. 
  }
  \label{fig:genmethod_qual}
\end{figure*}
\noindent\textbf{Component-wise Ablation.}
First, we conduct an ablation study on the Cityscapes$\rightarrow$ACDC adaptation scenario to analyze the impact of each component in our framework.
As shown in \Tref{tab:component}, when synthetic images are utilized naively without incorporating our key components (a), the performance degrades by 1.89\% compared to ULDA (42.47$\rightarrow$40.58).
Applying Domain Mix and Patch Style Transfer, which are specifically designed to reflect real-world characteristics and leverage the style variations of synthetic images((b),(c)), leads to performance improvements of 1.54\% and 2.23\%, respectively.
This demonstrates that simulating real-world style intensity variations helps the model better understand diverse styles. 
While both Domain Mix and Patch Style Transfer individually contribute to improved performance, leveraging all components yields the best score. This result highlights their complementary effects in the adaptation process.
\input{tables/imagenum}

\noindent\textbf{Impact of noise variance.}
\Fref{fig:noise_ablation} shows 
the hyperparameter analysis on the Gaussian noise $\boldsymbol\epsilon,\boldsymbol\epsilon^\prime$ with varying variance, added to Domain Mix and Patch Style Transfer.
We use $s_e=0.075$, which shows the best performance.
Note that our method shows superior performance compared to the baseline (ULDA) with the noise range of $s_e(0.075\sim0.1)$, showing the advantages of adding noise.

\noindent\textbf{The number of synthetic images.}
\Tref{tab:imagenum} illustrates the results of the ablation analysis on the number of synthetic images used during adaptation.
In most cases, even with an extremely limited number of images, our method consistently outperforms the state-of-the-art ULDA~\cite{yang2024unified}.
In our experiments, we utilized three images that yielded the highest performance.

\subsection{Analysis}
\noindent\textbf{Effect of generation methods.}
To select an appropriate synthetic image generation approach,
we conduct experiments using multiple approaches for generating images. First, the Direct T2I approach employs a fixed prompt, ``Driving at \{domain\}.'' to generate synthetic images.
Second, the Direct I2I approach involves selecting an arbitrary source image and directly applying the ``Driving at \{domain\}.'' prompt within a translation model.
Third, the GPT T2I method generates prompts for image generation by leveraging only textual information without using any image input.
As shown in ~\Tref{table:genmethod}, our generation method shows the best adaptation performance. 
Furthermore, qualitative results in~\Fref{fig:genmethod_qual} show a comparison of generated images using various generation methods. 
Compared to other methods, our results exhibit similar object composition—such as trees, buildings, roads, and sky—across translated images.
In the Direct T2I method, the consistency of objects appearing across different domains is relatively
low, and the generated images tend to focus on only one or two objects. Additionally, in the Direct I2I method, translating directly
from the source image often leads to image degradation (e.g., Night), making it difficult to effectively represent the desired
target domain style.
In GPT T2I, unlike Direct T2I, images are generated with long and detailed captions, which leads to a broader variety of objects appearing in the scenes. 
However, this approach results in the depiction of entirely different scenes across domains.
As a result, while the VLM-based method achieved the highest quantitative performance, it also demonstrated the most semantic consistency in qualitative evaluations.
We hypothesize that Domain Mix and Patch Style Transfer effectively model the variations that may arise when the object list is fixed.
\input{tables/generationmethod}

\input{tables/ULDA+ours}
\begin{figure}
    \centering
    \includegraphics[width=\linewidth]{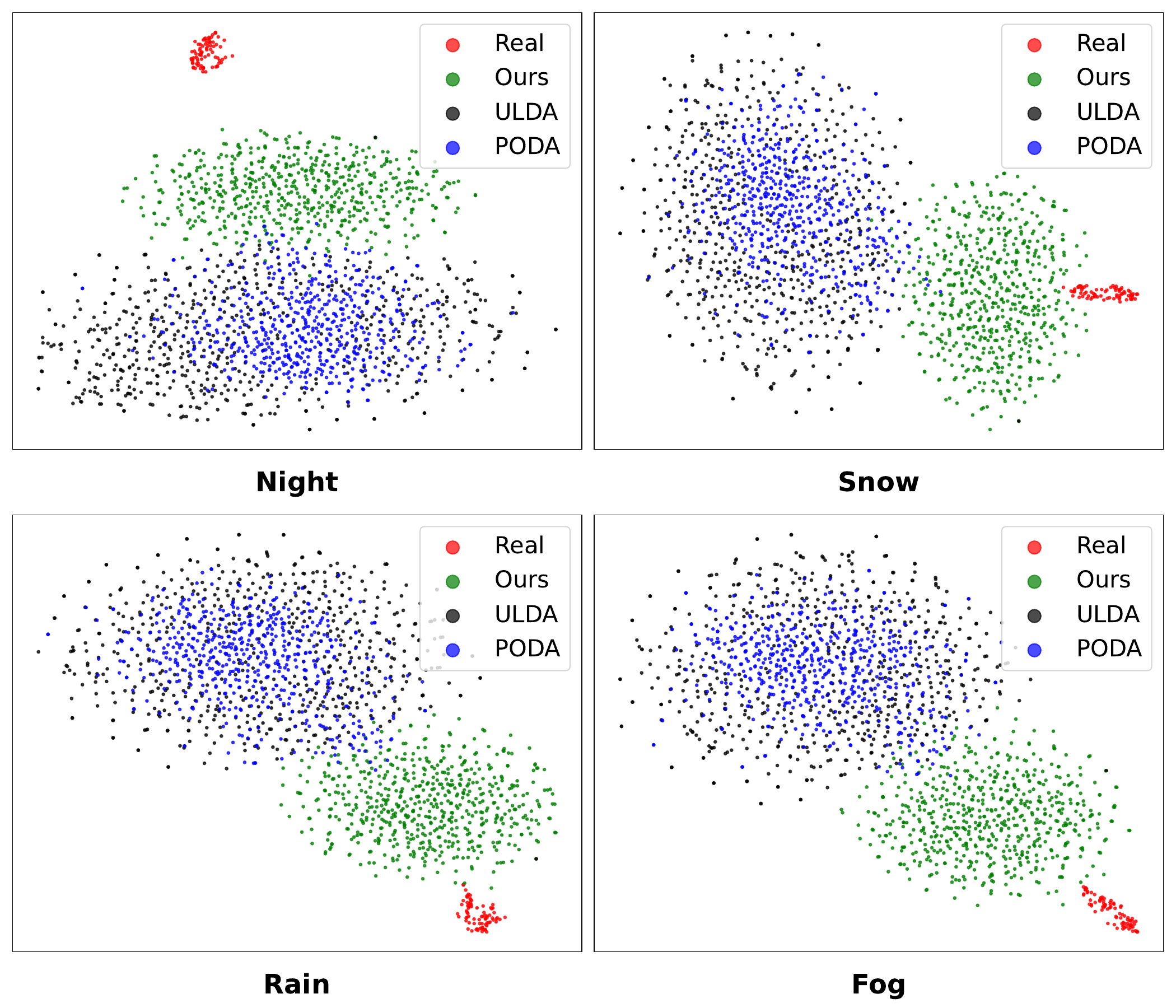}
    \caption{
    \textbf{The visualization results of the simulated target domain style obtained through each method.}
    \textcolor{red}{$\bullet$} represents the style features extracted from real target domain images.
    \textcolor{blue}{$\bullet$},\textcolor{black}{$\bullet$}, and \textcolor{green}{$\bullet$} denotes the style representations of stylized features produced by P{\O}DA\xspace~\cite{fahes2023poda}, ULDA~\cite{yang2024unified}, and $\model$, respectively.
    The features synthesized by our method have the closest distribution to the real dataset for all domains.
}
    \label{fig:tSNE}
\end{figure}
\noindent\textbf{Effectiveness of our components.}
In \Tref{tab:ULDA+ours}, we incorporate Domain Mix and Patch Style Transfer into existing methods~\cite{fahes2023poda,yang2024unified} to assess their effectiveness.
The experiments are conducted in the GTA5$\rightarrow$ACDC adaptation scenario.
During the adaptation process, we maintain the text-driven style optimization stage (Stage 1) of P{\O}DA\xspace and ULDA.
We then employ Domain Mix and Patch Style Transfer on the style features generated through the text-driven method.
As shown in the results, integrating Domain Mix and Patch Style Transfer into both P{\O}DA\xspace and ULDA consistently improves performance.
These results highlight that Domain Mix and Patch Style Transfer serve as efficient and effective feature augmentation techniques that do not require additional optimization steps. 
Notably, they exhibit strong adaptability across both image-based features and text-driven features, reinforcing their applicability to a wide range of ZSDA methodologies.

\noindent\textbf{Feature visualization.}
To evaluate the effectiveness of our method in simulating real-world styles, we perform a t-SNE analysis on the distribution of the simulated feature's style.
Specifically, we simulate target-like style features using various ZSDA methods, including $\model$, P{\O}DA\xspace, and ULDA, for visualization.
A style transfer algorithm is then applied to the source image features and the corresponding simulated target-like style features to generate target-like stylized features used in the analysis.
As shown in \Fref{fig:tSNE}, our method consistently narrows the distributional gap between the simulated and actual target domain style features across all domains.
These results demonstrate that our approach is capable of generating style features closely resembling those in real-world target domains.

\noindent\textbf{Adaptation efficiency.}
To evaluate the adaptation efficiency of each method, we measured the adaptation time three times and reported the average.
The official implementation of ULDA, referred to as ULDA\textsubscript{s}\footnote{\url{https://github.com/Yangsenqiao/ULDA}}, generates target-like style features using a sampling-based approach rather than utilizing the entire set of source images.
ULDA$^{*}$ is a re-implementation version that removes the sampling process and utilizes all source images.
All experiments were conducted using a single NVIDIA RTX 4090 GPU.
As shown in \Tref{tab:efficient}, the previous methods~\cite{fahes2023poda,yang2024unified} adaptation process is time-consuming due to the alignment step, and adaptation time heavily relies on the size of the source dataset.   
ULDA\textsubscript{s} reduces the absolute adaptation time by sampling strategy; however, it results in performance degradation in the Cityscapes$\rightarrow$ACDC adaptation scenario, with the mean mIoU decreasing from 42.47 to 41.66 and still remaining dependent on the size of the source dataset.
In contrast, $\model$ achieves significantly faster and better adaptation performance in both experimental settings.
This improvement stems from the $\model$ does not rely on costly text-image alignment processes.
Furthermore, unlike other methods, the $T_{GTA}/T_{CS}$ values of our approach are close to 1, which allows it to maintain a nearly constant adaptation efficiency regardless of the source dataset size.
\input{tables/efficient}

%% file: tables/main_experiment.tex
\begin{table}[t]
\caption{\textbf{Performance comparison of zero-shot domain adaptation in semantic segmentation.}
    Our method shows the best performance compared to the previous methods~\cite{kwon2022clipstyler,fahes2023poda,yang2024unified} across various scenarios.}
    \centering
    \label{tab:comparison}
    \resizebox{0.85\columnwidth}{!}{
    \begin{tabular}{lllc}
        \toprule
        {Source} & {Target eval.} & {Method} & {mIoU [\%]} \\
        \midrule
        
        \multirow{25}{*}{CS}
            & \multirow{5}{*}{ACDC Night}
                & Source-only & 18.31 \\
            &             & CLIPstyler & 21.38 \\
            &             & P{\O}DA\xspace  & 25.03 \\
            &             & ULDA       & 25.40 \\
            &             & $\model$       & \textbf{25.73}\tiny$\pm$0.21 \\
        \cmidrule(lr){2-4}
            & \multirow{5}{*}{ACDC Snow}
                & Source-only & 39.28 \\
            &             & CLIPstyler & 41.09 \\
            &             & P{\O}DA\xspace  & 43.90 \\
            &             & ULDA       & 46.00 \\
            &             & $\model$       & \textbf{46.48}\tiny$\pm$0.59 \\
        \cmidrule(lr){2-4}
            & \multirow{5}{*}{ACDC Rain}
                & Source-only & 38.20 \\
            &             & CLIPstyler & 37.17 \\
            &             & P{\O}DA\xspace  & 42.31 \\
            &             & ULDA       & 44.94 \\
            &             & $\model$       & \textbf{46.45}\tiny$\pm$0.21 \\
        \cmidrule(lr){2-4}
            & \multirow{5}{*}{ACDC Fog}
                & Source-only & 38.20 \\
            &             & CLIPstyler & 37.17 \\
            &             & P{\O}DA\xspace  & 51.54 \\
            &             & ULDA       & 53.55 \\
            &             & $\model$       & \textbf{53.84}\tiny$\pm$0.43 \\
        \cmidrule(lr){2-4}
            & \multirow{5}{*}{GTA5}
                & Source-only & 39.59 \\
            &             & CLIPstyler & 38.73 \\
            &             & P{\O}DA\xspace  & 40.77 \\
            &             & ULDA       & 42.91 \\
            &             & $\model$       & \textbf{44.45}\tiny$\pm$0.27 \\
        \midrule
        
        \multirow{5}{*}{GTA5}
            & \multirow{5}{*}{Cityscapes}
                & Source-only & 36.38 \\
            &             & CLIPstyler & 32.40 \\
            &             & P{\O}DA\xspace  & 40.02 \\
            &             & ULDA       & 41.73 \\
            &             & $\model$       & \textbf{41.98}\tiny$\pm$0.22 \\
        
        \bottomrule
    \end{tabular}
    }
    \label{tab:main}
\end{table}

%% file: tables/sand_fire.tex
\begin{table}[t]
\caption{\textbf{Performance comparison in the Fire and Sandstorm adaptation scenario.} 
$\model$ achieves state-of-the-art performance in the most challenging scenarios.}
\centering
\setlength\tabcolsep{6pt} 
\begin{adjustbox}{width=\linewidth,center=\linewidth} 
\begin{tabular}{l|cc|c }
\toprule

\multirow{3}{*}{Method}   & \multicolumn{2}{c|}{Scenarios} & \multirow{3}{*}{Mean-mIoU} \\
 & \multicolumn{1}{c}{CS $\rightarrow$ Fire} & \multicolumn{1}{c|}{CS $\rightarrow$ Sandstorm} &   \\ 
  &  mIoU[\%]  & mIoU[\%]  & \\ 
\midrule

Source-only & 10.08 & 19.58 & 14.83 \\ 
CLIPstyler~\cite{kwon2022clipstyler}  & 12.36 & 23.16 & 17.76\\ 
$\text{P{\O}DA\xspace}$~\cite{fahes2023poda} & 15.43 & 24.39 & 19.91\\ 
ULDA~\cite{yang2024unified}  & 19.52 & 25.72 & 22.62\\ 

$\model$ & \textbf{22.14} & \textbf{28.02} & \textbf{25.08}\\ 

\bottomrule
\end{tabular}
\end{adjustbox}
\label{tab:fire&sand}
\end{table}

%% file: tables/component_ablation.tex
\begin{table}[!tb]
\caption{{Ablation study of the key components of $\model$.} 
Patch ST and Ent-loss refer to Patch Style Transfer and weight CE loss. Our method with all the proposed components shows the best performance.}
\centering
\small 
\begin{adjustbox}{width=0.9\linewidth,center}
\setlength\tabcolsep{4pt} 
\renewcommand\arraystretch{1.1} 
\begin{tabular}{c|ccc c}
\toprule
Model &Patch ST  & Domain Mix & Ent-loss  & Mean-mIoU \\\midrule
    (a) &     &           &      & 40.58 \\ 
(b)&\checkmark  &           &      & 42.81\\
(c)&          & \checkmark &      & 42.12\\
(d) & \checkmark          & \checkmark &      & 42.85\\
Ours &\checkmark  &\checkmark &\checkmark  & \textbf{43.13}\\
\bottomrule
\end{tabular}
\end{adjustbox}
\label{tab:component}
\end{table}

%% file: tables/imagenum.tex
\begin{table}[]
\caption{Ablation study on the effect of the number of synthetic images.
Regardless of the number of images used, our approach consistently surpasses the performance of existing state-of-the-art scores across most settings.}
\centering
\resizebox{\columnwidth}{!}{
\begin{tabular}{c|c|ccccc}
\toprule
Number of Images                                                                     & ULDA~\cite{yang2024unified} & 1     & 3     & 5     & 10    & 100   \\ 
\midrule
\multicolumn{1}{c|}{\begin{tabular}[c]{@{}c@{}}CS to ACDC\\ (Mean-mIoU)\end{tabular}} &42.47 & 42.53 & $\mathbf{43.13}$ & 42.51 & 42.44 & 42.53 \\ 
\bottomrule
\end{tabular}
}
\label{tab:imagenum}
\end{table}

%% file: tables/generationmethod.tex
\begin{table}[]
\caption{{Ablation study on the synthetic image generation methods. {Leveraging both VLM-based prompt generation and image-to-image translation shows the best performance.}}}
\centering
\resizebox{\columnwidth}{!}{
\begin{tabular}{c|cccc}
\toprule
Generation Method                                                                     & Direct T2I     & GPT T2I     & Direct I2I     & \makecell{VLM T2I+I2I\\ \textbf{(Ours)}}      \\ 
\midrule
\multicolumn{1}{c|}{\begin{tabular}[c]{@{}c@{}}CS to ACDC\\ (Mean-mIoU)\end{tabular}} & 42.45 & 42.12 & 41.95 & $\mathbf{43.13}$ \\ 
\bottomrule
\end{tabular}
}
\label{table:genmethod}
\end{table}

%% file: tables/ULDA+ours.tex
\definecolor{deepgreen}{RGB}{0, 100, 0}
\begin{table*}[htb]
\caption{\textbf{Results of applying the $\model$ components to the existing ZSDA models, P{\O}DA\xspace~\cite{fahes2023poda} and ULDA~\cite{yang2024unified}.} 
We use the GTA5 dataset as the source domain and ACDC as the four target domains in this setting. 
Integrating $\model$ components into conventional methods generally leads to performance improvements.} 
\centering
\renewcommand{\arraystretch}{0.9}
\setlength\tabcolsep{11pt} 
\resizebox{0.8\textwidth}{!}{
\begin{tabular}{c|cccc|c }
\toprule

\multicolumn{1}{c|}{Scenarios}          & \multicolumn{1}{c|}{GTA $\rightarrow$ Night}    & \multicolumn{1}{c|}{GTA $\rightarrow$ Snow}     & \multicolumn{1}{c|}{GTA $\rightarrow$ Rain}  & \multicolumn{1}{c|}{GTA $\rightarrow$ Fog}    & \multirow{2}{*}{Mean-mIoU}  \\ 
Method & \multicolumn{1}{c|}{mIoU}  & \multicolumn{1}{c|}{mIoU}& \multicolumn{1}{c|}{mIoU}& \multicolumn{1}{c|}{mIoU}  & \\ 
\midrule
P{\O}DA\xspace~\cite{fahes2023poda} & 13.53 & 33.81 & 34.19 & 35.76 & 29.33 \\ 
P{\O}DA\xspace+$\model$  & \textbf{18.38} \textcolor{deepgreen}{(+4.85)} & \textbf{34.75} \textcolor{deepgreen}{(+0.94)} & \textbf{35.18} \textcolor{deepgreen}{(+0.66)} & \textbf{36.39} \textcolor{deepgreen}{(+0.53)} & \textbf{31.17} \textcolor{deepgreen}{(+1.84)}\\   
\midrule
ULDA~\cite{yang2024unified}  & 15.72 & 35.77 & \textbf{35.84} & {36.98} & 31.08 \\ 
ULDA+$\model$ & \textbf{17.30} \textcolor{deepgreen}{ (+1.58)} & \textbf{36.21} \textcolor{deepgreen}{ (+0.44)} & 35.77 \textcolor{red}{ (-0.07)} & \textbf{37.27} \textcolor{deepgreen}{ (+0.29)} & \textbf{31.64} \textcolor{deepgreen}{ (+0.56)} \\  

\bottomrule
\end{tabular}
}
\label{tab:ULDA+ours}
\end{table*}

%% file: tables/efficient.tex
\begin{table}[t]
\caption{\textbf{The efficiency of $\model$.}     Among the two experimental settings, CS$\rightarrow$ACDC utilizes 2,975 source images, while GTA$\rightarrow$CS leverages 12,382 source images. We report adaptation time ratios for existing methods relative to ours. The ratio of adaptation times ($T_{GTA}/T_{CS}$) indicates the relative adaptation time when employing the larger GTA5 source dataset compared to the smaller CS dataset. {Our method requires significantly less adaptation time compared to the state-of-the-art methods.}}
\centering
\resizebox{\linewidth}{1.3cm}{%
\begin{tabular}{l|c|c|c}


\toprule
\multirow{2}{*}{Method}  & \multicolumn{2}{c|}{Adaptation Time}    & \multirow{2}{*}{$T_{GTA}/T_{CS}$} \\ 
        & CS$\rightarrow$ACDC    & GTA$\rightarrow$CS   &                        \\ \midrule
PODA    & $8.97\times10^2~{(\times2.11)}$  & $2.72\times10^3 ~{(\times4.35)}$  & $3.03 $                  \\ 
ULDA\textsubscript{s} & $2.81\times10^3~{(\times6.11)}$  & $1.71\times10^4~{(\times27.32)}$  & $ 6.09  $                \\ 
 ULDA$^{*}$    & $5.06\times 10^3~{(\times 11.91)}$ & $1.28\times10^5~{(\times204.47)}$ & $25.30$                   \\ 
$\model$    & \(\mathbf{4.25 \times 10^2}\)   & \(\mathbf{6.26 \times 10^2}\)   & $\mathbf{1.47}$               \\ 
\bottomrule

\end{tabular}
}
\label{tab:efficient}
\end{table}

%% file: Sec/06_conclusion.tex
\renewcommand{\shortauthors}{Kim et al.}
\section{Conclusion}
In this paper, we propose $\model$, an efficient and effective ZSDA method.
Previous methods require aligning features to fixed textual descriptions using pre-trained CLIP text embeddings, which limits their capability to adequately and efficiently reflect style variations. 
Our model enables efficient adaptation by leveraging synthetic images, eliminating the need for a time-consuming alignment phase, and successfully reflecting the intensity of diverse global and local styles observed in real-world scenarios through Domain Mix and Patch Style Transfer.
Extensive experiments across various ZSDA scenarios demonstrate the effectiveness of our proposed method.
\label{sec:conclusion}